\documentclass[10pt,twocolumn,letterpaper]{article}

\usepackage[pagenumbers]{cvpr} 

\usepackage{multirow}
\usepackage[accsupp]{axessibility}
\usepackage{algorithm}
\usepackage{algorithmic}
\usepackage{graphicx}
\usepackage{amsmath}
\usepackage{amssymb}
\usepackage{booktabs}

\usepackage{adjustbox}
\usepackage{caption}
\usepackage{colortbl}
\usepackage{makecell}
\usepackage{float}
\usepackage{paralist}
\usepackage{color}
\usepackage{threeparttable}
\usepackage{lipsum}
\usepackage{url}

\usepackage[dvipsnames]{xcolor}

\definecolor{cvprblue}{rgb}{0.21,0.49,0.74}
\usepackage[pagebackref,breaklinks,colorlinks,citecolor=cvprblue]{hyperref}

\def\paperID{***} 
\def\confName{cvpr}
\def\confYear{1111}

\title{Learning Triangular Distribution in Visual World}

\author{
Ping Chen\dag$^{1}$, Xingpeng Zhang\dag$^{4}$,  Chengtao Zhou$^{1}$, Dichao Fan$^{1}$,  Peng Tu$^{3}$,  Le Zhang$^{1}$, Yanlin Qian$^{1,2}$\thanks{Corresponding author. \dag These authors contributed equally.}\\ 
    $^{1}$MicroBT Inc.
    $^{2}$Waseda University, IPS.
    $^{3}$RuqiMobility Inc.\quad\\
	$^{4}$School of Computer Science and Software Engineering, Southwest Petroleum University, Chengdu, China
}

\begin{document}
\maketitle
\begin{abstract}
Convolution neural network is successful in pervasive vision tasks, including label distribution learning, which usually takes the form of learning an injection from the non-linear visual features to the well-defined labels. However, how the discrepancy between features is mapped to the label discrepancy is ambient, and its correctness is not guaranteed.To address these problems, we study the mathematical connection between feature and its label, presenting a general and simple framework for label distribution learning. We propose a so-called Triangular Distribution Transform (TDT) to build an injective function between feature and label, guaranteeing that any symmetric feature discrepancy linearly reflects the difference between labels. The proposed TDT can be used as a plug-in in mainstream backbone networks to address different label distribution learning tasks. Experiments on Facial Age Recognition, Illumination Chromaticity Estimation, and Aesthetics assessment show that TDT achieves on-par or better results than the prior arts.  Code is available at \href{https://github.com/redcping/TDT}{https://github.com/redcping/TDT}.
\vspace{-1ex}
\end{abstract}

\begin{figure*}[h]
    \centerline{\includegraphics[width=0.9\textwidth]{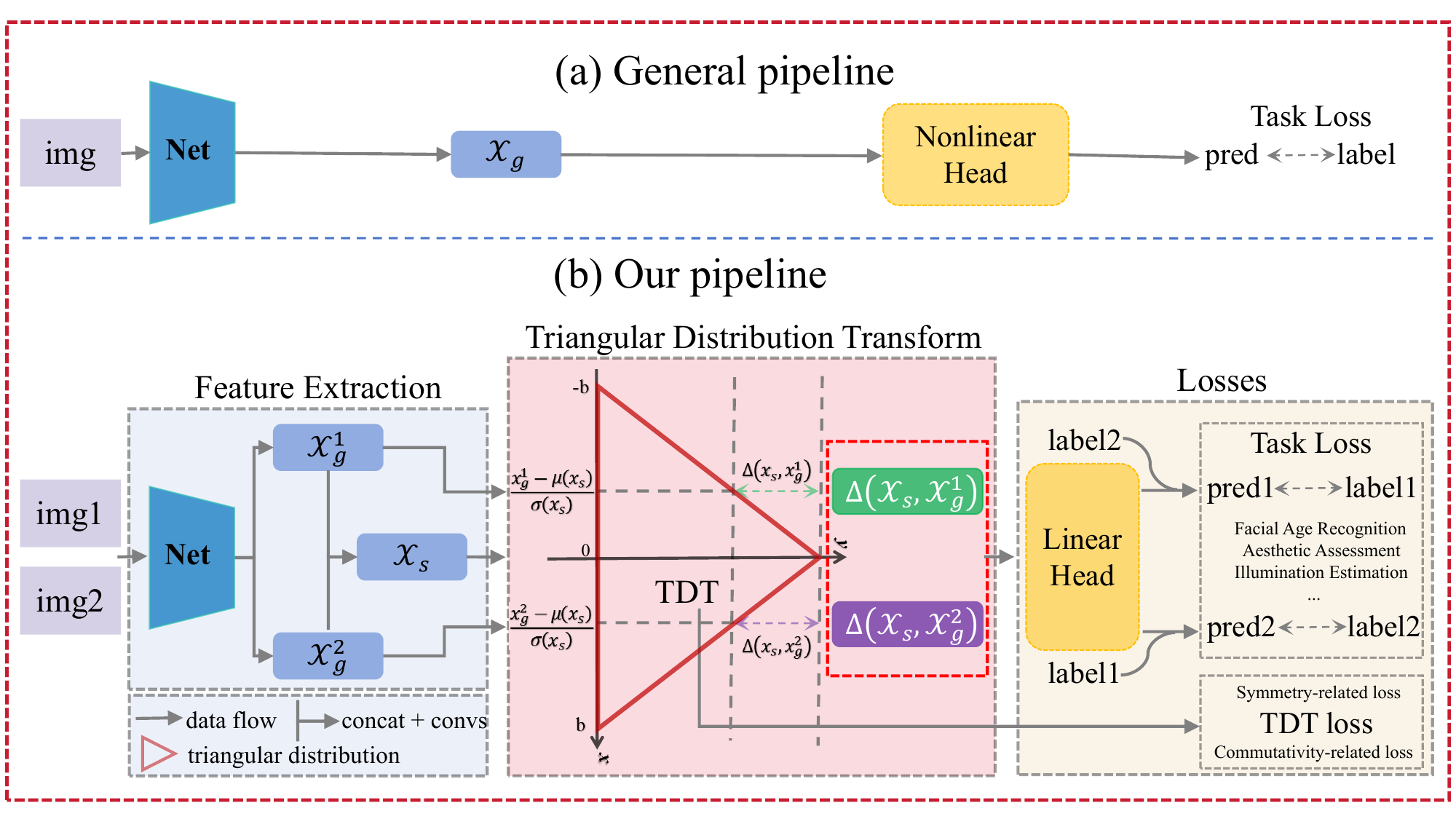}} 
	\centering
	\caption{The overall structure of our TDT. Our pipeline (b) diverges from the general pipeline (a) by incorporating a parameter-free TDT (Triangular Distribution Transform), 
 enabling converting the nonlinear feature to the one which vary ``linearly'' as per Eq.\ref{eq-F}.
 Consequently, a linear head module alone suffices to establish the mapping between image features and their respective labels, with clear explanation which is missing for a conventional network head. 
    For detailed information on the TDT loss, please refer to Figure.\ref{fig2}.}
	\label{fig1}
\end{figure*}
\section{Introduction}
\label{sec:intro}

Label distribution learning (LDL) utilizes the advantages of distribution to solve the task of quasi-continuous label or inner-correlation between labels \cite{Geng2013Deep, Shen2021Deep}. 
LDL assigns a distribution over label value to an instance, which can be obtained by fitting a Gaussian or Triangle distribution whose peak indicates the label and represents the relative importance of each label 
describing an instance \cite{Shen2021Deep}. 
Hence, LDL has an impact on many real-world applications, such as facial age estimation \cite{Geng2013Deep}, head-pose estimation \cite{Geng2014head, Geng2022head}, crowd counting \cite{Geng2015crowd}, zero-shot learning \cite{Huo2017zero}, facial beauty prediction \cite{Ren2017zero}, hierarchical classification \cite{Xu2019Hierarchical}, partial multi-label learning \cite{Xu2020Partial} and so on. 

When LDL was first proposed, it mainly used maximum entropy and Kullback-Leibler divergence to learn label distribution \cite{Geng2013Deep}. 
Then, a more efficient optimization method BFGS is proposed to replace the IIS method \cite{Geng2013Label}. The LDL is also widely combined with other algorithms, such as random forest \cite{shen2022nerp}, deep convolution neural network \cite{Yang2015Deep}, hashing method \cite{Zhang2022Hashing}, Bayesian \cite{Zheng2021Bayesian}, metric learning \cite{Zhou2022Facial} and so on. 
Although LDL has wide applications, 
it encounters some challenges, i.e. the feature space and label (solution) space are inhomogeneous \cite{shen2022nerp}. 
To be more specific, the feature space generated by the model is nonlinear, while the label information is gradually changing. 
In addition, the existing label distribution is mainly based on label information to construct the distribution, 
placing the label information for one instance over the whole label distribution.


Therefore, this article proposes a Triangular Distribution Transform (TDT) method, aiming to linearly map the transformed feature information to its corresponding label information 
concisely and efficiently.
Inspired by label distribution learning \cite{shen2022nerp}, 
we propose using symmetric triangular distributions to represent this symmetric linear transformation. 
Specifically, the high-dimensional features obtained by feature extraction networks will reflect the relationship between labels through symmetric triangular distributions. 
To better achieve this goal, we draw inspiration from the paradigm of comparative learning and supply two sets of images to the feature extraction network each time. One set is the prior knowledge that needs to be compared, and the other set is the training sample. Our method is more suitable for visual tasks with linearly continuous changing label information, such as age, aesthetics, lighting intensity, etc. Our TDT can be used as a plug-in in mainstream backbone networks. Therefore, our method has achieved excellent results on multiple visual tasks, such as facial age estimation, image aesthetics estimation, and illumination estimation.

The contribution of this article is summarized as follows:

\par\noindent$\bullet$ We analyze to lay the theoretical foundation for the Triangular Distribution Transform, enabling feature discrepancy to explain label difference.
\par\noindent$\bullet$ We show with the proposed symmetry-related loss and commutativity-related loss, TDT can be learnt by mainstream backbone networks.
\par\noindent$\bullet$ TDT outperforms other methods on age estimation, aesthetics estimation, and illumination estimation.

\section{Related works}
\label{sec:formatting}

\subsection{Label distribution learning}
The label distribution method learns the relevance between labels to reflect the relative importance of different labels \cite{Geng2010Facial, Geng2013Deep, He2017Data, Shen2021Deep}, which can also be seen as a special facial age classification method. Label distribution learning methods learn a label distribution that represents the relative importance of each label when describing an instance\cite{Geng2010Facial}. The label distribution covers a certain number of labels. Each label has its description degree, representing the degree to which each label describes the instance\cite{Su2019Soft}. The description degrees of all the labels sum up to 1 \cite{Su2019Soft}. Due to the advantages of label distributed learning, it has achieved very excellent performance in tasks such as facial age estimation\cite{Geng2013Deep}, head-pose estimation \cite{Geng2014head, Geng2022head}, crowd counting \cite{Geng2015crowd}, zero-shot learning\cite{Huo2017zero}, facial beauty prediction\cite{Ren2017zero}, hierarchical classification\cite{Xu2019Hierarchical}, and partial multi-label learning \cite{Xu2020Partial}. The label distribution learning method is very consistent with the potential law of big data. Nevertheless, acquiring distributional labels for thousands of face images itself is a non-trivial task\cite{Shen2021Deep}.

\subsection{Facial Age estimation}
The regression methods~\cite{Niu2016Ordinal}, classification methods \cite{Rothe2018Deep}, and ranking methods \cite{Chang2015A, Chen2017Using} for age estimation pay more attention to putting forward different research methods according to label information. The age regression methods consider labels as continuous numerical values. To handle the heterogeneous data, researchers also proposed hierarchical models \cite{Han2015Demographic} and the soft-margin mixture of regression \cite{Huang2017Soft}. And age classification regards labels as independent values\cite{Rothe2018Deep}. It regards each age as a separate category and ignores the similarity of the same person between different ages. While the ranking approaches treat labels as rank-order data and use multiple binary classifiers to determine the age rank in a facial image\cite{Chang2015A, Chen2017Using}. Besides,  some scholars also focus on the objective optimization function \cite{Pan2018Mean, Deng2021PML}. ML-loss \cite{Pan2018Mean} proposed mean-variance loss for robust age estimation via distribution learning. Deng et al. \cite{Deng2021PML} proposed progressive margin loss (PML) for long-tailed age classification.
These methods gradually consider that aging is a slow and continuous process, which also means that the processing of label information is significant. 

\subsection{Aesthetic Assessment}
An aesthetic assessment task refers to evaluating the visual beauty and artistic value of given image. 
Early studies rely on handcrafted features \cite{dhar2011high,nishiyama2011aesthetic,marchesotti2011assessing}, which ignore the spatial features and semantics in assessing aesthetics.
In data-driven learning-based methods, NIMA\cite{talebi2018nima} uses Earth Mover's Distance to optimize aesthetic distribution prediction. 
A-lamp\cite{ma2017lamp} and MP$_{ada}$\cite{sheng2018attention} both achieve good results with a multi-patch approach. 
Hierarchical Layout-aware Graph Convolution Network (HGCN)\cite{she2021hierarchical} captures layout information. TANet \cite{he2022rethinking} adaptively learns aesthetic prediction rules based on identified themes, using Mean Squared Error as the metric. Transformer \cite{he2023thinking} assign attention levels to color spaces, enabling segmentation learning. Yi \etal \cite{yi2023towards} effectively combine feature style and general aesthetic information using AdaIN \cite{huang2017arbitrary} and self-supervised pre-training for accurate aesthetic assessment.

\subsection{Illumination Estimation}

Illumination Estimation is often dubbed as Auto White Balance verbally and aims at measuring the normalized illumination vector given at least one single image.
There exists a bunch of traditional methods that are easy to implement but easily fail due to the over-optimistic assumption, for example, 
White Patch \cite{brainard1986analysis}, General Gray World \cite{barnard2002comparison}, Gray Edge \cite{van2007edge}, Shades-of-Gray \cite{finlayson2004shades}, LSRS \cite{gao2014eccv}, PCA \cite{cheng2014illuminant} and Grayness Index~\cite{Qian_2019_CVPR}, \etc.
Relying on labeled data, a bunch of learning-based methods with tunable inner weight \eg \cite{gijsenij2010generalized,barron2015convolutional, chakrabarti2015color,chakrabarti2012color,finlayson2013corrected,gehler2008bayesian,gijsenij2011color,hu2017cvpr,joze2014exemplar,shi2016eccv,yanlin2017iccv} leads the leaderboard by a large gap generally.  All aforementioned methods output a single illumination vector in a deterministic way. A few works deal with label distribution learning;  Egor \etal \cite{ershov2023physically} proposed an efficient illumination distribution estimation method; FFCC~\cite{Barron2017FFCC} from Barron outputs a unique illumination vector which can be modified as a distribution. 

\begin{figure*}[h]
    \centerline{\includegraphics[width=0.95\textwidth]{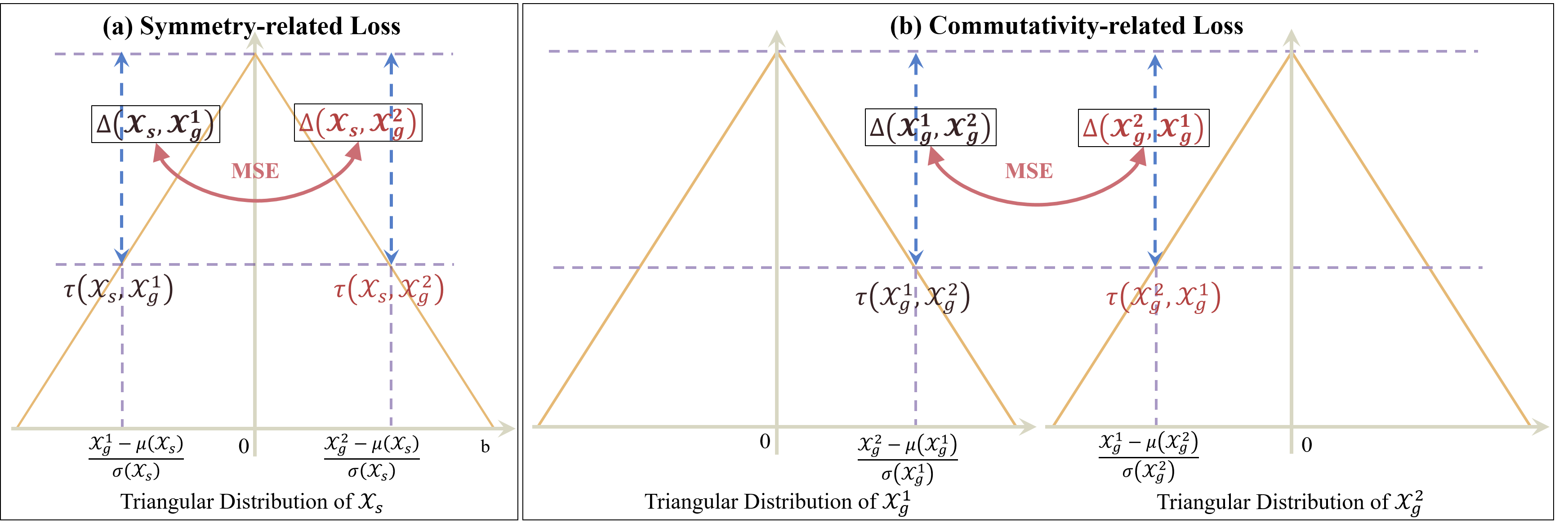}} 
	\centering
	\caption{
 TDT is learned relying on the commutativity-related loss and symmetry-related loss, while the latter plays the primary role. The feature difference, associated with the symmetry-related loss, is used for result prediction. MSE is the mean square error.}
	\label{fig2}
\end{figure*}
\section{Proposed Method}

The pipeline of proposed Triangular distribution learning is illustrated in Fig.\ref{fig1}.
Fig.\ref{fig1}(a) depicts the general pipeline for CNN tasks, where non-linear head modules are employed to map non-linear features to labels due to the linear independence of image features. These non-linear head modules can include modules that combine convolution with non-linear activation functions, Gaussian Mixture Model, and others.
In Fig.\ref{fig1}(b), we propose a novel pipeline that differs from the general pipeline. We introduce a parameter-free TDT (Triangular Distribution Transform) after the non-linear image features, allowing for the possibility of transforming linearly independent image features into linearly correlated ones. As a result, we only need to utilize a linear head module, such as a fully connected (FC) module, in combination with the relevant loss function to achieve linear correlation learning among the features. Additionally, we employ a contrastive learning-like approach to guide the predictions of unknown samples based on known samples.

\subsection{Triangular Distribution Transform on Latent Feature}

With some ambiguity, vision regression problems can be classified into linear-label problems and nonlinear-label ones.
The mapping from the original information modality (for example 2D images, text, voice, video \textit{etc.}) to the corresponding labels is usually non-linear and hard to model in a fixed rule.
Take facial age recognition as an example, age label varies linearly while the visual feature of the facial image usually does not work in the same way.
The majority of works on distribution learning mainly focus on learning a mapping from visual features to the label in a nonlinear way, such as the deep net, with its interpretability remaining in suspense. 
In this article, we discuss the relationship between the feature and its label and present a so-called Triangular Distribution  Transform (TDT) to rigidly connect the feature and its label, so that the connection is injective. 

Starting with an input $X \in \mathbb{R}^{N\times 3 \times H \times W}$ ($N$ is the batch size, $3, H, W$ denotes the channel, height, and width respectively) and it label $Y$,  we extract feature by some net module $g$(parameterized by $\Theta$) as follows: 
\begin{equation}\label{eq-1}
	\mathcal{X}_{g} = g(\mathcal{X}, \Theta)
\end{equation} 
where $\mathcal{X}_{g} \in \mathbb{R}^{N\times C \times h \times w}$ is a C-channel feature extracted from some backbone nets (\textit{e.g.}, ResNet18). 
The mean and standard deviation are calculated along the $h$ and $w$ dimensions,  referred to as $\mu(\mathcal{X}_g) \in \mathbb{R}^{N\times C \times 1 \times 1}$ and $\sigma(\mathcal{X}_g) \in \mathbb{R}^{N\times C \times 1 \times 1}$ respectively.


Based on the Central Limit Theorem, we assume $\mathcal{X}_g$ follows Gaussian distribution and denote the probability density function (PDF) as $\phi(\mathcal{X}_g)$.
Given features for two samples ($\mathcal{X}_g^1$ and $\mathcal{X}_g^2$),  we define their ''feature difference'' based on the Gaussian distribution of $\mathcal{X}_g^1$ as:
\begin{align}\label{eq-delta0}
\small
	\Delta(\mathcal{X}_g^1 ,\mathcal{X}_g^2) &  = \phi(\mathcal{X}_g^1) - \phi(\mathcal{X}_g^2) \\
   & \propto \mathcal{N}_{0,1}(\frac{\mathcal{X}_g^1 - \mu(\mathcal{X}_g^1)}{\sigma(\mathcal{X}_g^1)}) - \mathcal{N}_{0,1}(\frac{\mathcal{X}_g^2 - \mu(\mathcal{X}_g^1)}{\sigma(\mathcal{X}_g^1)}) \nonumber
\end{align}

The last part of Eq.\ref{eq-delta0} is the normalized feature difference using the standard Gaussian distribution $\mathcal{N}_{0,1}$. We give our first assumption:

\textbf{Assumption 1:} There exists a function transforming $\mathcal{X}_g$ to $\mathcal{X}_g'$, whose change linearly \textit{w.r.t.} the label.

To better fit Assumption 1, considering $\Delta(\mathcal{X}_g^1,\mathcal{X}_g^2)$ does not correlate linearly with the label, we approximate the Gaussian distribution in Eq.\ref{eq-delta0} using a symmetric Triangular distribution $\tau(\cdot | b)$ following \cite{Scherer2003triangular}, formulated as:
\begin{align}
\tiny
	\Delta_{}(\mathcal{X}_g^1 ,\mathcal{X}_g^2) & \approx  \Delta_\tau{}(\mathcal{X}_g^1 ,\mathcal{X}_g^2) \\
    & =  \tau(\frac{\mathcal{X}_g^1 - \mu(\mathcal{X}_g^1)}{\sigma(\mathcal{X}_g^1)} |b) - \tau(\frac{\mathcal{X}_g^2 - \mu(\mathcal{X}_g^1)}{\sigma(\mathcal{X}_g^1)} |b),  \nonumber \\
 	\tau(s|b) &=
	\begin{cases}
		\frac{1}{b}\cdot(1-\frac{\lvert s \lvert}{b}), &-b \leq s\leq b \\
		0, &otherwise
	\end{cases}
 \label{eq-F}
 \end{align}
where the scalar $b$ controls the approximation error and is set to $\sqrt{6}$ ($b$ is obtained via the method of moments in \cite{Scherer2003triangular}).

In contrast to Eq.\ref{eq-delta0}, Eq.\ref{eq-F} has the property of linearly describing the corresponding label difference. 
Then we can search a linear function $f(\Delta(\mathcal{X}_g^1,\mathcal{X}_g^2))$ that transforming the feature difference to the label difference $|Y^1-Y^2|$, in a linear, symmetric and commutative way\footnote{ 
Linearity: given $\Delta_\tau(\mathcal{X}_g^1,\mathcal{X}_g^2) = \Delta_\tau(\mathcal{X}_g^2,\mathcal{X}_g^3)$, we have $Y^1-Y^2 = Y^2-Y^3$.\\ 
Symmetry:  $\Delta_\tau(\mathcal{X}_g^1,\mathcal{X}_g^1+\delta \mathcal{X}_g) = \Delta_\tau(\mathcal{X}_g^1,\mathcal{X}_g^1 - \delta \mathcal{X}_g)$.\\
Commutativity: $\Delta_\tau(\mathcal{X}_g^1,\mathcal{X}_g^2) = \Delta_\tau(\mathcal{X}_g^2,\mathcal{X}_g^1)$.}.

\subsection{Optimization towards Triangular Distribution Transform and Vision Tasks }

The learned feature does not follow Triangular distribution unless we optimize the net towards it. Then we state the optimization loss towards TDT and related task-specific loss. 
The optimization loss of TDT, shown in Fig.\ref{fig2}, aims to gradually transform the non-linear correlation among image features into a linear correlation through label-guided learning.

\textbf{Loss for Triangular Distribution}

Given two sets of nonlinear features $\mathcal{X}_g^1$ and $\mathcal{X}_g^2$ extracted from backbone network, 
we use multiple convolution layers to learn to make a feature that follows symmetric Triangular distribution, formulated as $\mathcal{X}_s = Convs(Concat(\mathcal{X}_g^1, \mathcal{X}_g^2), \theta)$, where the concatenation happens in the batch dimension. We calculate the mean and standard deviation separately for $\mathcal{X}_s$, $\mathcal{X}_g^1$ and $\mathcal{X}_g^2$ along the spatial dimensions. Via Eq.\ref{eq-F}, we get $\Delta(\mathcal{X}_s, \mathcal{X}_g^1)$ and $\Delta(\mathcal{X}_s, \mathcal{X}_g^2)$. 

To guide the net to reach a state that $\mathcal{X}_g^1$ and $\mathcal{X}_g^2$ are symmetrical around the ``center'' $\mathcal{X}_s$, a symmetry-related loss is introduced: 
\begin{align}
\small{
	\textit{L}_S = \| \Delta(\mathcal{X}_s, \mathcal{X}_g^1) - \Delta(\mathcal{X}_s, \mathcal{X}_g^2)  \|_2,
 }
 \label{eq-symloss}
\end{align}
where the operator $\|\cdot\|_2$ is a L2 loss. Analogously, the commutativity-related loss is designed as: 
\begin{align}
\small{
	\textit{L}_M = \| \Delta(\mathcal{X}_g^1, \mathcal{X}_g^2) - \Delta(\mathcal{X}_g^2, \mathcal{X}_g^1)  \|_2,
 }
 \label{eq-analoss}
\end{align}

Then, to facilitate the learning of the linear function $f(\Delta)$, we adopt a comparative approach and design a relevant supervised loss:
\begin{align}
	L_{P} & = \|\delta(Y) -  2\cdot f(\Delta(\mathcal{X}_s, \mathcal{X}_g^1)\cdot sgn(\mathcal{X}_s,\mathcal{X}_g^1)) \|_1  \nonumber \\
     & + \| \delta(Y) + 2\cdot f(\Delta(\mathcal{X}_s, \mathcal{X}_g^2)\cdot sgn(\mathcal{X}_s,\mathcal{X}_g^2))\|_1,
     \label{eq-Lp}
\end{align}
Here, $\|\cdot\|_1$ refers to smooth L1 operator, $\delta(Y) = Y^2-Y^1$, $sgn(\mathcal{X}_s,\mathcal{X}_g)$ represents the sign of $\Delta(\mathcal{X}_s, \mathcal{X}_g)$. When $\mathcal{X}_g$ in the distribution of $\mathcal{X}_s$ has a cumulative distribution function (CDF) less than 0.5, $sgn(\mathcal{X}_s,\mathcal{X}_g)$ is considered positive, and vice versa. The first part of  Eq.\ref{eq-Lp} represents using samples labeled as $Y^2$ to predict samples labeled as $Y^1$, and the other term follows the same principle.

Therefore, the final loss function for learning the triangular distribution can be formulated as follows:
\begin{align}\label{eq-Lt}
\small{
    L_{T} =  L_S + L_M + L_P,
    }
\end{align}

\begin{table*}
	\renewcommand{\arraystretch}{1}
	\caption{Experiments setting for three tasks evaluated.}
	\label{table_ES}
	\centering
	\begin{tabular}{cccc}
		\hline
		Task & Baseline& Dataset  & Input size \\
		\hline
		Facial Age Recognition & DAA(Resnet18)\cite{Chen2023DAA} & MegaAge-Asian\cite{Zhang2017Quantifying} & $96\times96$ \\
		Aesthetic Assessment &SAAN(Resnet50+VGG19) \cite{yi2023towards} & TAD66K\cite{he2022rethinking} & $256\times256$\\
		Illumination Estimation & FC$^4$(FC, SqueezeNet)\cite{Hu2017FC4}& Reprocessed Color Checker \cite{gehler2008bayesian,shi2010re}& $512\times512$\\
		\hline
	\end{tabular}
\vspace{-2ex}
\end{table*}

\textbf{Loss for Vision Tasks}

To quickly validate the effectiveness of TDT, we have selected three common visual tasks: facial age recognition, aesthetic assessment, and illumination estimation. Below are the descriptions and loss definitions for these three tasks.

\textbf{\textit{Facial Age Recognition:}} Age variation can be considered linear,  and we can utilize TDT to draw the relationship between visual feature differences among different images and the corresponding age variation. Therefore, we can directly employ Eq.\ref{eq-Lt} for optimizing age estimation.

\textbf{\textit{Aesthetic Assessment:}}  The aesthetic assessment variation in scores can be considered linear, similar to the optimization for the age estimation task, we can also choose Eq.\ref{eq-Lt} as the overall optimization. In this case, the label corresponds to the aesthetic score.

\textbf{\textit{Illumination Estimation:}} In illumination estimation, a key step of automatic white balance, we consider the variation of the illumination to be approximately linear. Additionally, to better capture the differences in illumination angles, we combine Eq.\ref{eq-Lt} and the angle error \cite{Hu2017FC4} as the overall optimization function  $L_{ill} =  L_T + \frac{180}{\pi}\cdot arccos(p\cdot p^*)$,
where the $p$ and $p^*$ represent the normalized estimation and ground truth of illumination color, respectively.

\begin{algorithm}
\caption{The training process of our method}
\label{alg1}
\begin{algorithmic}[1]
\STATE \textbf{Input:} Sample set $X$ and corresponding labels $Y$
\STATE \textbf{Output:} Loss
\STATE \textbf{Prior Set Selection:} Randomly select $N$ samples $X_2$ from the training set with labels $Y_2$. 
\FOR{each sample $X_1$ in forward pass}
    \STATE \textbf{\# Feature fusion and distribution generation}
    \STATE $\mathcal{X}_g^1, \mathcal{X}_g^2 \gets g(X_1, \Theta), g(X_2, \Theta)$  \# Extract features
    \STATE $N, C, h, w \gets \mathcal{X}_g^2.\text{shape}$ \# $C$ is distribution number
    \STATE $\mathcal{X}_g^1 \gets \mathcal{X}_g^1.repeat(N, 1, 1, 1)$
    \STATE $\mathcal{X}_s \gets conv(concat(\mathcal{X}_g^1, \mathcal{X}_g^2, 1), C)$ 
    \STATE \textbf{\# Perform TDT operation on $\mathcal{X}_g^1$ based on $\mathcal{X}_s$}
    \STATE $\mu(\mathcal{X}_s), \sigma(\mathcal{X}_s) \gets calc\_mean\_std(\mathcal{X}_s)$ 
    \STATE $\Delta_{}(\mathcal{X}_s ,\mathcal{X}_g^1) \gets \tau(\frac{\mathcal{X}_s - \mu(\mathcal{X}_s)}{\sigma(\mathcal{X}_s)} |b) - \tau(\frac{\mathcal{X}_g^1 - \mu(\mathcal{X}_s)}{\sigma(\mathcal{X}_s)} |b)$
    \STATE \textbf{\# Feature-to-Label Difference Mapping}
    \STATE $sgn(\mathcal{X}_s,\mathcal{X}_g^1) \gets (-1)^{cdf(\mathcal{X}_g^1) > 0.5}$ 
    \STATE $\delta Y_1 \gets f(\Delta(\mathcal{X}_s, \mathcal{X}_g^1) \cdot sgn(\mathcal{X}_s, \mathcal{X}_g^1))$ \# via Eq.7
    \STATE \hspace{\algorithmicindent} where $f(\cdot) = fc(gap(\cdot))$ 
    \STATE \textbf{Optimize TDT, including $L_S$, $L_M$, and $L_P$}
\ENDFOR
\end{algorithmic}
\end{algorithm}

\subsection{TDT Algorithm Process}
To be more clear, we provide pseudocode \ref{alg1}, in 5 parts.

\textbf{Prior Samples Selection:} 
Prior samples are several selected training samples, providing comparison with those samples used for training/testing.
Features for the prior samples can be pre-computed and integrated into the model.

\textbf{Distribution generation:} We obtain $\mathcal{X}_s$ and form its high-dimensional Normal Distribution using $\mu(\mathcal{X}_s)$ and $\sigma(\mathcal{X}_s)$. After standardizing and approximating, we get a zero-symmetric high-dimensional Triangular Distribution to linearly represent feature differences.

\textbf{TDT Operation:} 
It calculates the PDF differences between $\mathcal{X}_s$ and $\mathcal{X}_g$ based on the distribution of $\mathcal{X}_s$ meet
$|\Delta(\mathcal{X}_g^1,\mathcal{X}_g^2)| = 2 \cdot |\Delta(\mathcal{X}_s,\mathcal{X}_g^1)| = 2 \cdot |\Delta(\mathcal{X}_s,\mathcal{X}_g^2)|$. 
Remarkably, TDT is entirely formulaic, parameter-free, and fit any net outputting feature maps.

\textbf{Difference Mapping:} Base on distribution of $\mathcal{X}_s$, $\mathcal{X}_s$ generally has a higher PDF than $\mathcal{X}_g$,
 meaning $\Delta{}(\mathcal{X}_s ,\mathcal{X}_g) = pdf(\mathcal{X}_s) - pdf(\mathcal{X}_g) \geq 0$. We infer feature difference signs from $cdf(\mathcal{X}_g) > 0.5$ and linearly map them to label difference via global average pooling and a linear layer.

\textbf{Optimization:} For commutativity loss, we utilize the mean and std of $\mathcal{X}_g^1$ and $\mathcal{X}_g^2$. For symmetry and supervisory losses, those of $\mathcal{X}_s$ are computed. 

\section{Experiments}
\subsection{Notes and Implementation Details}
We perform TDT validation on the baseline in three tasks, whose configuration details are shown in Table \ref{table_ES}.

\textbf{Train and Test notes:}
As observed in pseudocode \ref{alg1}, our TDT aims to guide the learning of unknown samples using prior samples, resembling contrastive learning. Therefore, for each task experiment, we randomly select a prior set from the training dataset. During training and testing, each batch comprises both the prior set samples used to obtain $\mathcal{X}_g^2$ and the unknown samples used to obtain $\mathcal{X}_g^1$. Each unknown sample undergoes TDT operations with all prior samples to make predictions, and the average of these predictions is considered as the final prediction. To reduce feature extraction time during testing, we encapsulate all the features from the prior set into the model parameters, resulting in an optimized model with accelerated inference. 

\textbf{Common Setting}: 
For all experiments, we used the Adam optimizer, where the weight decay and the momentum were set to $0.0005$ and $0.9$, respectively. The initial learning rate was set to $0.001$ and changed according to cosine learning rate decay. We trained our model using PyTorch on a cluster of 8 RTX 3090 GPUs. In an online manner, we augment all images with random horizontal flipping, scaling, rotation, and translation.



\textbf{Facial Age Recognition}: 
We utilize the \emph{FG-Net} \cite{Panis2016Overview} dataset, which comprises 1002 facial images from 82 subjects, spanning an age range from 0 to 69. We follow the setup described in the papers \cite{Chen2023DAA, Li2019BridgeNet, Deng2021PML}, employing leave-one-person-out (LOPO) cross-validation. We report the average performance over 82 splits using the Mean Absolute Error (MAE) as the evaluation metric.


\begin{table}[t]
	\renewcommand{\arraystretch}{1.0}
	\caption{CA  on MegaAge-Asian.}
	\label{table_Mega}
	\centering
 \begin{adjustbox}{width=0.46\textwidth}
	\begin{tabular}{ccccc}
		\hline
		Methods & Pre-trained  & CA(3) & CA(5) & CA(7) \\
		\hline
		Posterior \cite{Zhang2017Quantifying} & IMDB-WIKI & 62.08 & 80.43 & 90.42\\
		MobileNet \cite{Sandler2018mobile} & IMDB-WIKI & 44.0 & 60.6 & - \\
		DenseNet \cite{Yang2018SSR}  & IMDB-WIKI & 51.7 & 69.4 & - \\
		SSR-Net \cite{Yang2018SSR}  &  IMDB-WIKI  & 54.9 &74.1 & -\\
        UVA \cite{li2019uva} &-& 60.47 &79.95& 90.44 \\
        LRN(ResNet10) \cite{Li2020Deep} & IMDB-WIKI & 62.86 & 81.47 &  91.34 \\
        LRN(ResNet18) \cite{Li2020Deep} & IMDB-WIKI & 64.45 & 82.95 & 91.98 \\
		VGG16(norm) \cite{Zhao2021Distilling} & ImageNet, IMDB-WIKI, AFAD & 65.58 & 83.01 & 89.17 \\
		PVP+VGG16 \cite{Zhao2021Distilling} & ImageNet, IMDB-WIKI, AFAD & \textbf{72.65} & \textbf{87.24} & {\color{blue}93.16} \\
		DAA(single channel) \cite{Chen2023DAA} & - & 67.97 & 84.06&92.40 \\
        DAA(multi-channel) \cite{Chen2023DAA} & - & 68.29& 84.84& 92.47 \\
        DAA \cite{Chen2023DAA} & - & 68.82 & 84.89 & 92.70 \\
        \hline
		\textbf{TDT (ours)} & - & {\color{blue}69.60} & {\color{blue}85.42} & \textbf{93.26} \\
		\hline
	\end{tabular}
 \end{adjustbox}
\end{table}

Additionally, we also use the \emph{MegaAge-Asian} \cite{Zhang2017Quantifying} Dataset, consisting of 40,000 age-labeled samples spanning 0 to 70 years and 3,945 images for testing. For this dataset, we choose cumulative accuracy (CA) \cite{Zhang2017Quantifying} as the evaluation metric, defined as $CA(n) = \frac{K_{n}}{K}\times100$, 
where $K$ is total number of testing images and $K_{n}$ is the number whose absolute errors are smaller than $n$.

We adopt the official implementation of DAA\cite{Chen2023DAA} as the baseline for this task while replacing its DAA mapping with our TDT. The output size of $\mathcal{X}_g$ is set to $3\times3$.

In this task, 256 images are randomly picked from the training set serves as the prior set for all experiments.
\begin{table}[t]
	\renewcommand{\arraystretch}{1.0}
	\caption{MAEs on FG-Net dataset.}
	\label{table_FG}
	\centering
  \begin{adjustbox}{width=0.2\textwidth}
	\begin{tabular}{ccc}
		\hline
		Methods & MAE & Year\\
		\hline
		DEX\cite{Rothe2015DEX} & 3.09 & 2015\\
        MV\cite{Pan2018Mean} & 2.68 & 2018 \\
        C3AE\cite{Zhang2019C3AE} & 2.95 & 2019\\
		DRFs\cite{Shen2021Deep} & 3.85 & 2021\\
		PML\cite{Deng2021PML} & {\color{blue}2.16} & 2021\\
        DAA \cite{Chen2023DAA} & 2.19 & 2023\\
		\hline
		\textbf{TDT(ours)} & \textbf{2.12} & -\\
		\hline
	\end{tabular}
 \end{adjustbox}
 \vspace{-1ex}
\end{table}

\textbf{Aesthetic Assessment}: 
We use \emph{TAD66K (Theme and Aesthetics Dataset with 66K images)} Dataset \cite{he2022rethinking}, a large-scale aesthetic quality assessment database with  47 themes. Images belonging to each theme are annotated independently, with each image containing a minimum of 1200 valid annotations.
The aesthetic score of each image is treated as the label. Following TANet\cite{he2022rethinking}, the evaluation metric is the Mean Squared Error (MSE).
We adopt the same train-test split setting.
For the prior set, we select samples by choosing 2 random samples for each score ranging from 0 to 10 with interval 0.1  for each theme.
We employ the official implementation of SAAN\cite{yi2023towards} as the baseline for this task while removing the final average pooling and BN layers. Additionally, we incorporate extra convolutions to
obtain  $\mathcal{X}_g$ with size of $8\times8$.

\textbf{Illumination Estimation}: 
We use \emph{reprocessed Color Checker} \cite{gehler2008bayesian,shi2010re} dataset, one of the most widely adopted datasets in illumination estimation. 
Following FC$^4$ \cite{Hu2017FC4}, the evaluation metric is the Recovery Angular Error. To facilitate better comparisons, we use the linear fully connected version of FC$^4$-net as the baseline, instead of the non-linear weight pooling version.
With additional convolutional layers, we obtain $\mathcal{X}_g \in \mathbb{R}^{N\times C \times 8 \times 8}$.
When comparing with the FC$^4$ model, we randomly select 128 samples from the training set as the prior set.  However, when comparing with the method proposed by Tang  \etal\cite{Tang2022TLCC} that utilizes the extended external sRGB datasets, we choose same images for scene classification as the prior set, but only 1280 samples. 


Following previous AWB works\cite{Hu2017FC4,gehler2008bayesian,shi2010re}, 3-fold cross-validation is used.  
Standard metrics (mean, median, tri-mean of all the errors, the mean of the lowest 25\%, and the mean of the highest 25\% of errors, the 95th percentile error) are reported in terms of angular error in degrees.


\begin{table}[t]
	\renewcommand{\arraystretch}{1}
	\caption{Results on Aesthetic Benchmark TAD66K.}
	\label{table_Aes}
  \resizebox{0.48\textwidth}{!}{
	\centering
	\begin{tabular}{cccc}
		\hline
		method & Pub. & Basic & MSE \\
		\hline
        RAPID \cite{lu2014rapid} & ACMMM2014 & incorporate heterogeneous & 0.0200 \\
        PAM\cite{ren2017personalized}	& ICCV2017	& residual-based, active learning & 0.0200 \\
        ALamp\cite{ma2017lamp} & CVPR2017 & layout-aware, multi-patch & 0.0190 \\
        NIMA \cite{talebi2018nima}& TIP2018 & predict distribution & 0.0210 \\
        MPada \cite{sheng2018attention} & ACMM2018 & attention, multi-patch & 0.0220 \\
        MLSP \cite{hosu2019effective}& CVPR2019 & staged training,multi-level features & 0.0190 \\
        UIAA \cite{zeng2019unified} & TIP2019	& unified probabilistic formulation	& 0.0210 \\
        HGCN \cite{she2021hierarchical} & CVPR2021 & graph convolution networks & 0.0200 \\
        TANet \cite{he2022rethinking} &IJCAI2022 & attention, adaptive features & \textbf{0.0161} \\
        SAAN \cite{yi2023towards}& CVPR2023	& AdaIN,self-supervised Pretraining & 0.0185 \\
        \hline
        \textbf{TDT(ours)} & - & Triangular Distribution & {\color{blue} 0.0172}\\
		
		\hline
	\end{tabular}
 }
 \vspace{-2ex}
\end{table}

\subsection{Results and Analysis}

\noindent \textbf{Facial Age Estimation}

The quantitative comparison of a list of top-performing methods on the MegaAge-Asian Dataset is shown in Table \ref{table_Mega}. 
From the table, we can find that even stocked with pre-training on large-scale datasets, methods like \cite{Zhao2021Distilling, Li2020Deep} are inferior to ours.
Compared with SSR-Net, our improvement is over 8\%. PVP \cite{Zhao2021Distilling} gets higher CA(3) and CA(5) indexes, due to the massive pretraining on ImageNet\cite{Deng2009ImageNet}, IMDB-WIKI\cite{Rothe2018Deep}, and AFAD\cite{Niu2016Ordinal}. DAA\cite{Chen2023DAA}, published in 2023, shares the same setting with our proposed method, while TDT obtains better scores in all CA indexes.

To better prove the effectiveness of TDT,  we also profile TDT on FG-Net \cite{Panis2016Overview},  reported in Table~\ref{table_FG}. Using MAE as the metric, TDT leads the board with the smallest MAE, surpassing the same-setting competitor DAA.

\noindent \textbf{Aesthetic Assessment.}

In Table \ref{table_Aes}, we list the MSE scores for the selected and close-related aesthetic works.  In an overall manner, TDT ranks second with an MSE score of $0.0172$.  This improvement simply originates from the deployment of the TDT plugin over SAAN \cite{yi2023towards}, whose MSE is $0.0185$, again showing the advantage of the label discriminative ability of TDT.


\begin{table}
	\renewcommand{\arraystretch}{1}
	\caption{Results on AWB Dataset Color Checker Dataset.}
	\label{table_AWB}
	\centering
 \resizebox{0.48\textwidth}{!}{
	\begin{tabular}{ccccccc}
		\hline
		models & mean & med. & tri. & best25\% & worst 25\% & 95th pct. \\
		\hline
        FFCC \cite{Barron2017FFCC} & 1.80 & 0.95 & 1.18 & {\color{blue}0.27} & 4.65 & - \\
        FC$^4$(Weighted) \cite{Hu2017FC4} & 1.65 & 1.18 & 1.27 &0.38 & 3.78 & 4.73 \\
        FC$^4$\cite{Hu2017FC4}	&1.84 &	1.27&1.39&0.46&4.20&5.46\\
        AlexNet FC4 \cite{Hu2017FC4} & 1.77 & 1.11 & 1.29 &0.34 & 3.78 & 4.29 \\    
		
        Multi-Hypothesis \cite{Hern2020Multi} & 2.10 & 1.32 & 1.53 & 0.36 & 5.10 & - \\
        IGTN \cite{Xu2020End} & 1.58 & {\color{blue}0.92} & - &  0.28 & 3.70 & - \\
        MDLCC \cite{Xiao2020Multi} & 1.58 & 0.95 & 1.11 & 0.37 & 3.77 & - \\
        TLCC+sRGB \cite{Tang2022TLCC} & {\color{blue}1.51} & 0.98 & 1.07 & 0.33 & \textbf{3.52} & - \\
		\hline
        \textbf{TDT+FC$^4$} & 1.64 & 1.12	&1.25&0.41&3.80& \textbf{4.53}\\
        \textbf{TDT+FC$^4$+sRGB} & \textbf{1.46} & \textbf{0.85} & \textbf{1.05} & \textbf{0.26} & {\color{blue}3.61} & {\color{blue}4.61} \\
        \hline
	\end{tabular}
 }
\end{table}
\noindent \textbf{Illumination Estimation.}

In Table \ref{table_AWB}, we report the statistics of the predicted angular error for a set of AWB algorithms on the Color Checker Dataset. 
On metrics like mean, and median values, the TDT based on FC$^4$ outperforms other state-of-the-art methods, such as MDLCC \cite{Xiao2020Multi} and TLCC \cite{Tang2022TLCC}.
On top of FC$^4$, TDT also boosts the performance by a noticeable gap. 

\begin{figure*}[t]
	\centerline{\includegraphics[width=0.9\linewidth]{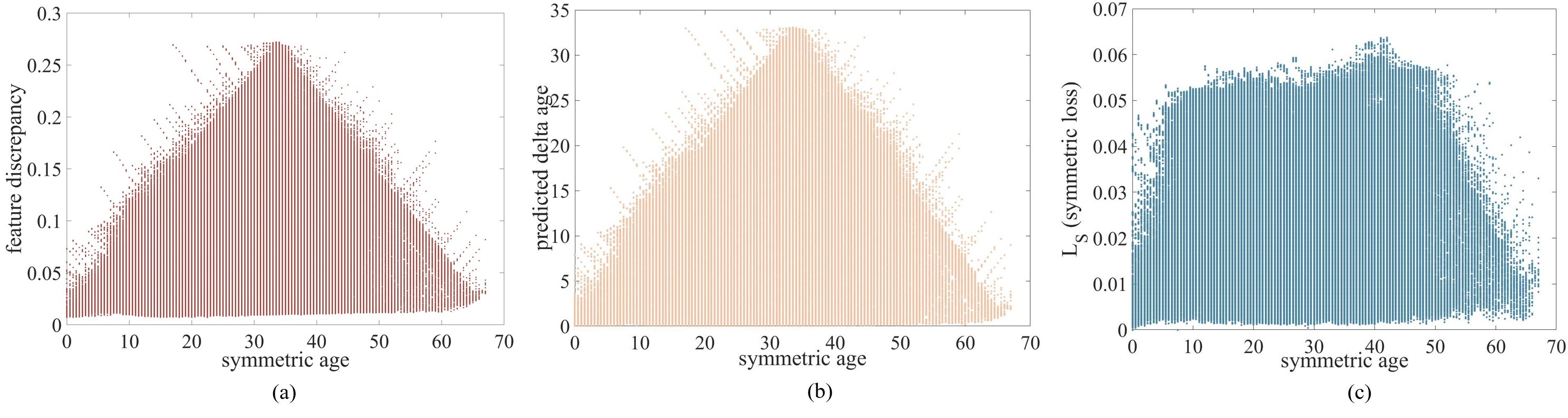}} %
	\centering
	\caption{Relationship between symmetric age with the feature discrepancy, predicted delta age, and symmetric loss. (a) feature discrepancy; (b) predicted delta age; (c) symmetric loss}
	\label{fig4}
\vspace{-1ex}
\end{figure*}

\subsection{Ablation Study}

\begin{table}[t]
	\renewcommand{\arraystretch}{1}
	\caption{Loss ablation analysis on MegaAge-Asian. $\checkmark$ and $-$ indicate whether to add this loss to the final loss.}
	\label{table_Loss}
	\centering
   \resizebox{0.34\textwidth}{!}{
	\begin{tabular}{ccc|ccc}
		\hline
		$L_{S}$& $L_{M}$ & $L_{P}$ &  CA(3) & CA(5) & CA(7) \\
		\hline
		- & \checkmark  & \checkmark & 67.74 & 84.33 & 92.51 \\
		\checkmark & - &  \checkmark & 68.43 & 84.62 & 92.60 \\
		\checkmark & \checkmark  & \checkmark & \textbf{69.60} & \textbf{85.42} & \textbf{93.26} \\
		\hline
	\end{tabular}
 }
\vspace{-2ex}
\end{table}

We conduct an ablation study on the loss choice, the number of distributions and prior samples.

\noindent \textbf{Loss function} 

It is vital to study the impact of different losses in Eq.\ref{eq-Lt} and the results are presented in Table \ref{table_Loss}.  We observe that removing any loss incurs a decrement in the overall performance. 
And symmetry loss brings the most improvement.



We design a toy experiment to further analyze the loss function Eq.\ref{eq-Lp}.  The age loss $L_P$ is a supervisory loss due to the label is needed for computing $L_P$, while the symmetry-related loss $L_S$ and the commutativity-related loss $L_M$ are unsupervisory losses. Switching off all nonsupervisory loss, as seen in the top two rows of Table \ref{table_L}, we see a clear CA score drop, indicating the advantage brought by the nonsupervisory loss. 

\begin{figure*}[t]
	\centerline{\includegraphics[width=0.9\linewidth]{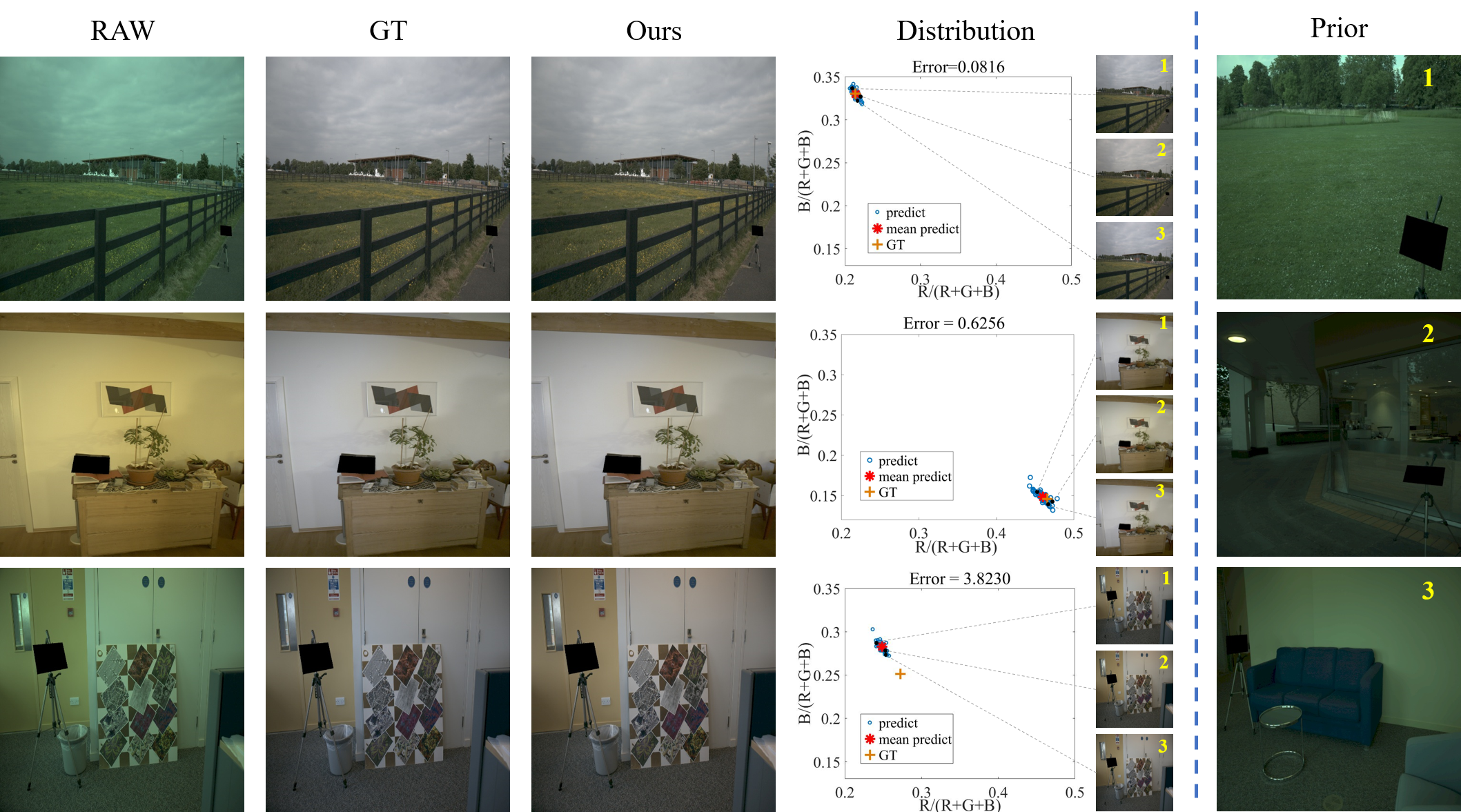}} %
	\centering
	\caption{
 Qualitative comparison of the final color correction result and those from individual prior samples. In the fourth column, the predictions from different priors (their color corrected images are given in right insets) are shown clustered around the ground truth location, with limited variance. 
 In the rightmost column, prior samples are given.
 }
\label{figILL}
\vspace{-1ex}
\end{figure*}

\begin{figure}
	\centerline{\includegraphics[width=0.9\linewidth]{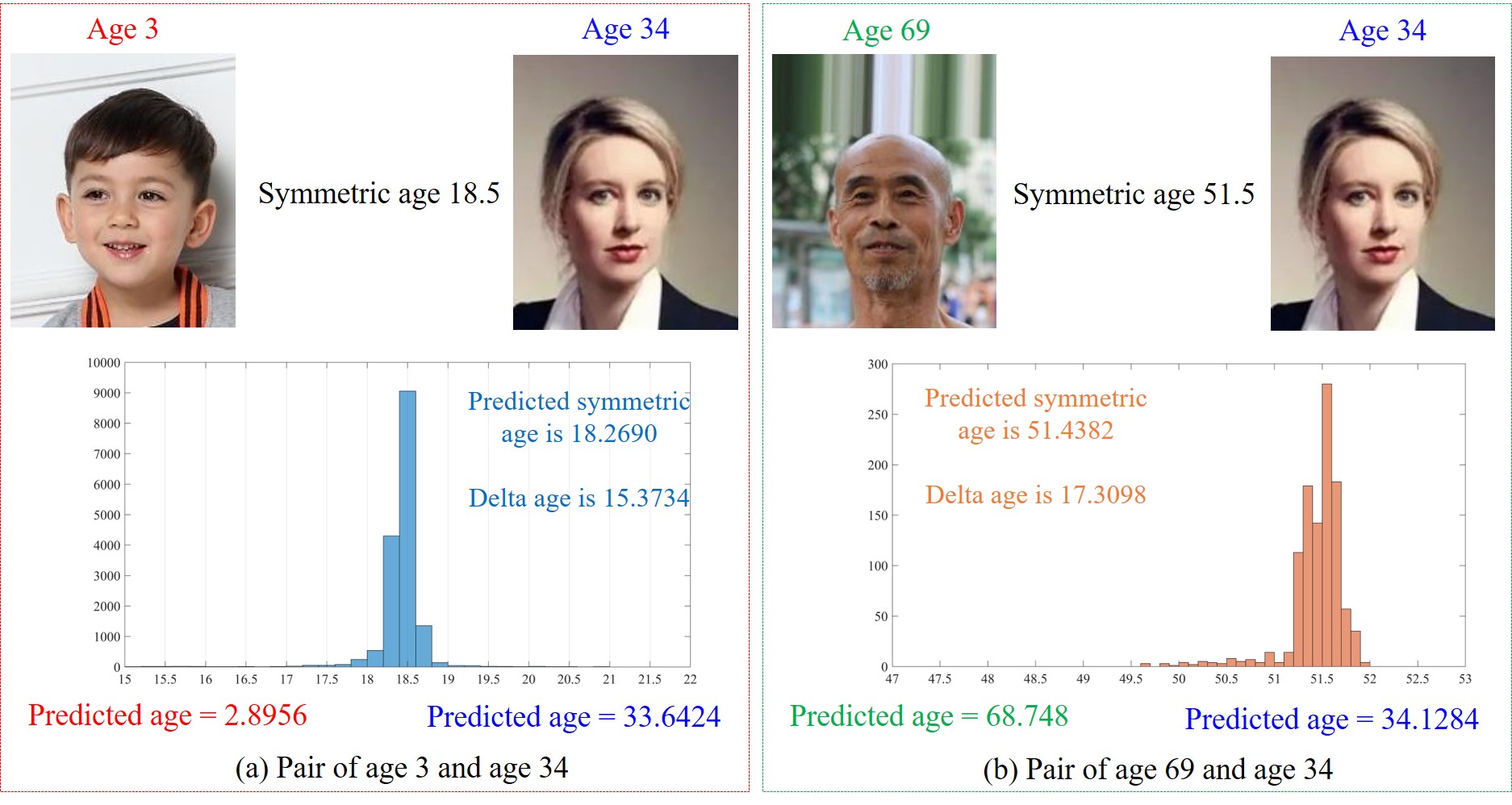}} %
	\centering
	\caption{
 Study on the symmetry learnt in $\mathcal{X}_s$.  
 For any pair of age-$a$ image from test set and age-$b$ image from the prior set, we draw a histogram of the resulted  $\mathcal{X}_s$, which in fact gather around the symmetry axis $(a+b)/2$, \wrt the histogram figure.
 }
\label{fig5}
\vspace{-4ex}
\end{figure}


If we cut off the access to unsupervisory loss for part1 (in other words, for only 75\% data, the unsupersiory loss is calculated), as reported in the bottom half of Table~\ref{table_L}, we claim: with increasing unlabeled data, the performance is massively improved, which again verifies the functionality of the symmetry-related and commutativity-related loss.

\noindent\textbf{Distribution and Prior Number}

Besides the loss selection, the number of distributions is a vital factor.
On the same MegaAge-Asian Dataset, we test the number of distributions from the set ${32, 64, 128, 256}$.
Table \ref{table_D} shows that in general, the number of distributions has a minor effect on the final results and when feature is set $128$-dimensional the performance is optimal.


The selected prior samples work like expert voting, hence we exploit the factor of the prior sample number. Table \ref{table_D} proves that as the number of samples increases, the credibility of the results improves, yet the enhancement in credibility gradually diminishes.


\begin{table}
	\renewcommand{\arraystretch}{1}
	\caption{
 Loss analysis on MegaAge-Asian. 
 The whole dataset is divided into part1 and part2, with a ratio of 25\% and 75\%.
 }
	\label{table_L}
	\centering
   \resizebox{0.4\textwidth}{!}{
	\begin{tabular}{c|c|c|c|c}
  		\hline
		$L_{S},L_{M}$& $L_{P}$& label accessed to &  CA(3) & CA(5) \\
		\hline
		No & all & all & 68.26 & 84.26  \\
		Yes  & all &  all & \textbf{69.60} & \textbf{85.42} \\
		\hline
		part2 & part2 & part2 & 60.23 & 78.54  \\
		all & part2 & part2 & \textbf{66.59} & \textbf{83.62}  \\
		\hline
	\end{tabular}
 }
\end{table}

\begin{table}[t]
	\renewcommand{\arraystretch}{1}
	\caption{Experiments on feature distributions and prior samples.}
	\label{table_D}
	\centering
   \resizebox{0.45\textwidth}{!}{
	\begin{tabular}{ccccc}
		\hline
		types & numbers & CA(3) & CA(5) & CA(7) \\
		\hline
		\multirow{4}{*}{feature distributions} & 256 & 68.77 & 84.82 & 92.70 \\
		& 128 & \textbf{69.60} & \textbf{85.42} & \textbf{93.26} \\
		& 64 & 68.39 & 84.59 & 92.85 \\
		& 32 & 67.80 & 85.04 & 92.90 \\
		\hline
        \multirow{3}{*}{prior samples} & 256 & \textbf{69.60} & \textbf{85.42} & \textbf{93.26} \\
		& 128 & 68.73 & 84.94 & 92.80 \\
		& 64  & 67.34 & 83.42 & 91.38 \\
  \hline 
	\end{tabular}
 }
\vspace{-3ex}
\end{table}


\subsection{Visualization of what TDT learns}


Fig.\ref{fig4} shows that: once the training is completed, we see feature difference approximates a triangular distribution function on varying symmetric age. The same property can be found for the predicted delta age (Inset (b)) in Fig.\ref{fig4}).  
This results from TDT learning. 
From the inset (c) in Fig.\ref{fig4}, we observe that the symmetry loss range is almost the same, meeting the feature symmetry assumption.





Fig.\ref{figILL} shows some qualitative comparison on illumination estimation task.  Starting from raw image,  different prior sample gives a close-to-groundtruth predicted white point, averaged to form the final robust white point.  For the 3rd row, the prediction slightly shifts away from the groundtruth, which we guess it is due to multi-illumination.

As shown in Fig.\ref{fig5}, we study the performance of TDT learning the symmetry axis location.
For any pair of facial images, for example an age-$a$ image from test set and an age-$b$ image from the prior set,  thus the symmetry axis should locate at $(a+b)/2$.
With $(a+b)/2$ fixed, we alter the testing image and  prior image,  and draw a histogram of the resulted  $\mathcal{X}_s$, which in fact gather around the symmetry axis.
For pairs of \{age-$3$, age-$34$\}, the mean of predicted symmetric age is $18.27$, close to the symmetric age $18.50$.

\section{Conclusion}
   A learning framework based on the Triangular Distribution Transform is proposed in this paper. It connects the nonlinear feature difference and the corresponding label difference in a verified linear, symmetric, and commutativity manner.  
   This transform can be used as a portable plug-in for vision regression tasks, \eg, we verify its application on three vision tasks. 
   On facial age estimation, aesthetic assessment, and illumination estimation, the proposed TDT obtains on-par or even better performance than the prior arts, without much modification on the affiliated backbone. 
   In the future, we will explore the application of TDT in a wider context, for example shape and pose estimation. 

\section{Acknowledgements}
This work is supported by the Natural Science Starting Project of SWPU (No.2022QHZ023), the Sichuan Scientific Innovation Fund (No.2022JDRC0009),  the Sichuan
Provincial Department of Science and Technology Project (No. 2022NSFSC0283) and the Key Research and Development Project of Sichuan Provincial Department of Science and Technology (No.2023YFG0129).

{
    \small
    \bibliographystyle{ieeenat_fullname}
    \bibliography{main}
}


\end{document}